\documentclass{article}

\usepackage[final, nonatbib]{nips_2017}

\usepackage[utf8]{inputenc} 
\usepackage[T1]{fontenc}    
\usepackage{hyperref}       
\usepackage{url}            
\usepackage{booktabs}       
\usepackage{amsfonts}       
\usepackage{nicefrac}       
\usepackage{microtype}      

\pdfoutput=1

\usepackage{amscd}
\usepackage{amsmath}
\usepackage{amssymb}
\usepackage{amsthm}
\usepackage{hyperref}
\usepackage{graphicx}
\usepackage{epstopdf}
\usepackage{latexsym}
\usepackage{verbatim}
\usepackage{booktabs}
\usepackage[labelformat=simple]{subcaption}

\usepackage{tikz}
\usetikzlibrary{arrows}
\input epsf.tex

\theoremstyle{definition}

\theoremstyle{remark}

\newcounter{tempthm}
\newcounter{tempsec}

\newcommand{\Rmnonneg}{\mathbb{R}_{\geq 0}}

\newcommand{\pobs}{p^{\text{obs}}}
\newcommand{\phatobs}{\hat{p}^{\text{obs}}}

\newcommand{\eqdef}{:=}

\makeatletter
\@ifpackagewith{nips_2017}{final}{
\newcommand{\ack}{
\subsubsection*{Acknowledgements}
The author thanks Paris Smaragdis, Johannes Traa, Th\'{e}ophane Weber, and David Wingate for their helpful suggestions regarding this work.  
}
\newcommand{\codeurl}{\url{http://arxiv.org/src/1411.5010/anc}}}{\newcommand{\ack}{}\newcommand{\codeurl}{available at ANONYMIZED}}

\title{Nonnegative Tensor Factorization for Directional Blind Audio Source Separation}

\author{Noah D. Stein \\
Analog Devices $\vert$ Lyric Labs: Cambridge, MA $02139$ \\
        \href{mailto:noah.stein@analog.com}{\nolinkurl{noah.stein@analog.com}}
}

\begin{document}

\maketitle

\begin{abstract}
We augment the nonnegative matrix factorization method for audio source separation with cues about directionality of sound propagation.  This improves separation quality greatly and removes the need for training data, with only a twofold increase in run time.  This is the first method which can exploit directional information from microphone arrays much smaller than the wavelength of sound, working both in simulation and in practice on millimeter-scale microphone arrays.
\end{abstract}

\section{Introduction}
Nonnegative matrix factorization (NMF) has proven to be an effective method for audio source separation \cite{rs:lvdsscss}.  We guide NMF to identify discrete sources by providing cues: direction of arrival (DOA) estimates computed from a microphone array for each time-frequency bin.  We form a (potentially sparse) frequency $\times$ time $\times$ direction tensor $X$ indicating the distribution of energy in the soundscape and jointly solve for all the sources by finding tensors $B$ (direction distribution per source), $W$ (spectral dictionary per source), and $H$ (time activations per dictionary element) to fit
\begin{equation*}
X(f,t,d) \approx \sum_{s,z} B(d,s)W(f,z,s)H(t,z,s),
\end{equation*}
where $s$ and $z$ index sources and dictionary subcomponents thereof.  The crucial fact that $B$ does not depend on $z$ forces the portions of the spectrogram explained by the dictionary for a given source to be collocated in space: if $B$ depended on $z$ in addition to $s$ like $W$ and $H$ do, then we could swap dictionary components between sources without affecting the sum and the model would have no power to isolate and separate cohesive sources of sound.  Advantages of our approach include:
\begin{itemize}
\item Perceived separation quality -- better than NMF;
\item No supervision -- no clean audio examples needed;
\item Low overhead -- computation on the same order as NMF;
\item Suitability for small arrays -- usable DOA estimates can be obtained from arrays which are much smaller than the wavelengths of audible sounds and for which beamforming fails.  We have tested this on millimeter-scale arrays; see Section~\ref{sec:experiments}.
\end{itemize}

We focus on small arrays to enable applications where industrial design constraints make it expensive to bring distant signals together for processing. Closer microphones make more integration possible, lowering cost and allowing for devices like smart watches whose sizes limit spacing. In real world use cases training data is often not available, especially for interferers, hence the desire for an unsupervised approach. In the end the perceived quality is better than the supervised version with longer but reasonable running time.

Previous work does not address the close microphone case; the conventional wisdom is that microphones separated by much less than the half wavelength required for beamforming provide no usable directional information. Linear unmixing methods like ICA \cite{s:bscmfd} fail because close spacing makes the mixing matrices ill-conditioned. DUET \cite{yr:bssmtfm} does not apply because there are essentially no amplitude differences between the microphone signals at this scale; a phase-only version relies too heavily on mixture-corrupted phase values and gives poor quality. Our method achieves good quality with necessarily poor direction information by enforcing the extra structure of an NMF decomposition for each source. NMF and variants are common for audio source separation, but here we extend them to solve a problem previously assumed hopeless.

Several approaches to extending NMF-like techniques to multichannel audio appear in the literature.  FitzGerald, Cranitch, and Coyle apply a form of nonnegative tensor factorization (NTF) to the frequency $\times$ times $\times$ number of channels tensor produced by stacking the spectrograms for all channels \cite{fcc:ntfsss}.  This approach has two drawbacks.  First, each dictionary element in the decomposition has its own gain to each channel, so a post-NTF clustering phase is needed to group dictionary elements into sources.  Second, it relies on level differences between channels to distinguish sources, so it only applies to widely spaced microphones or artificial mixtures.

Ozerov and F\'{e}votte address the first drawback by constraining the factorization so the gains from source to channel are shared between the dictionary elements which comprise each source \cite{of:mnmfcmass}.  Lee, Park, and Sung use a similar factorization and confront the second drawback by stacking spectrograms from beamformers in place of raw channels \cite{lps:bdmnmfass}.

While independent from our work, \cite{lps:bdmnmfass} can be viewed as an instance of the general method we present in Section~\ref{sec:ntfmodel} below, but the former still requires microphone spacing wide enough for beamforming.  Furthermore the computational resources required are proportional to the number of beamformers used, so for good spatial resolution the cost may be high.  In \cite{lps:bdmnmfass} all experiments use beamformers pointed north, south, east, and west, which may provide poor separation when the sources of interest are not axis-aligned.  

Also related is the paper \cite{tsws:dnmfjsls} on Directional NMF, which considers factoring the steered response power (SRP), a measure of the energy detected by a beamformer as a function of direction, time and frequency.  This three-dimensional tensor is flattened into a $\#\{\text{discretized directions}\}\times\#\{\text{spectrogram bins}\}$ matrix, to which NMF is applied with inner dimension equal to the number of sources sought.  Again this approach suffers in the face of closely-spaced microphones; furthermore it does not model any of the structure expected within reasonable audio spectrograms.  We compare our method experimentally to a variant of this in Section~\ref{sec:experiments}.


This paper is organized as follows.  Section~\ref{sec:background} covers basic NMF and a simple DOA estimator.  Our main contribution is introduced in Section~\ref{sec:ntfmodel} and efficient implementation is explored in Sections~\ref{sec:ntfmultupdate} and~\ref{sec:sparsedir}.  We show experimental results in Section~\ref{sec:experiments} and close with conclusions in Section~\ref{sec:conclusions}.

\section{Background}
\label{sec:background}
The material in this section is standard in audio source separation and could likely be skipped by an expert.  We present NMF here in equivalent probabilistic rather than matrix language \cite{dlp:ebnmfplsi} to make it easy to extend to the new algorithm we introduce in Section~\ref{sec:ntf} and ease comparison of algorithmic steps, convergence proofs, and run time.  This also gives an opportunity to make the seemingly new observation that Gibbs' inequality provides the optimizer in the M step of the EM algorithm for NMF.  Doing so we avoid Lagrange multiplier computations, which in any case cannot tell a maximizer from a minimizer.

\subsection{Nonnegative Matrix Factorization (NMF)}
\label{sec:nmfderivation}
NMF is a technique to factor an elementwise nonnegative matrix $X\in\Rmnonneg^{F\times T}$ as $X \approx WH$ with $W\in\Rmnonneg^{F\times Z}$ and $H\in\Rmnonneg^{Z\times T}$.  The fixed inner dimension $Z\ll F,T$ controls model complexity.  This technique is often applied to a time-frequency representation $X$ (e.g.\ a magnitude spectrogram) of an audio signal \cite{rs:lvdsscss}.

We use probabilistic language, so we normalize and in place of $X$ take a given probability distribution $\pobs(f,t)$ to decompose as $\pobs(f,t) \approx \sum_z q(f,t,z)$, where $q$ is factored into either of the equivalent forms
\begin{equation}
\label{eq:nmfdecomp}
q(f,t,z) \eqdef q(f,z)q(t \mid z) = q(f \mid z)q(t,z),
\end{equation}
as in Figure~\ref{fig:zft}. The values of $z$ index a dictionary of prototype spectra $q(f \mid z)$ which combine according to the time activations $q(t, z)$.  We seek to maximize the \textbf{cross entropy}
\begin{equation*}
\begin{split}
\alpha(q)\eqdef \sum_{f,t}\pobs(f,t)\log q(f,t) = \sum_{f,t}\pobs(f,t)\log\sum_z q(f,t,z)
\end{split}
\end{equation*}
for simplicity of exposition, though other objectives may provide better performance \cite{kfs:ocfmpnmfbssmi}.

\tikzset{observed node/.style={circle,minimum size=0.8cm,fill=gray!30,draw,font=\sffamily\large\bfseries},
unobserved node/.style={observed node,fill=white}
}

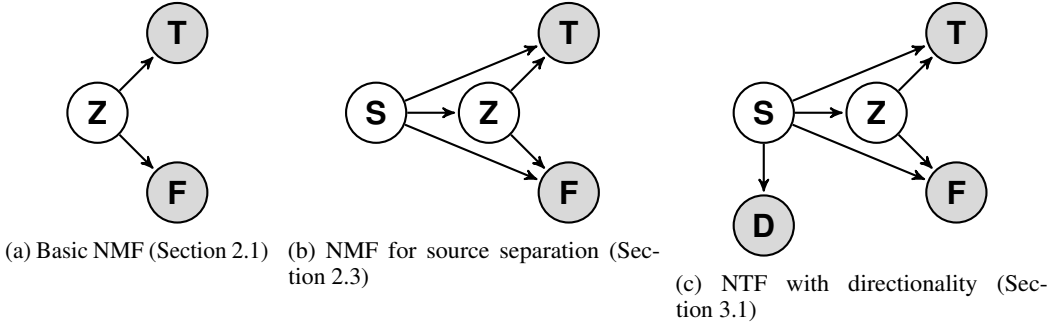
\begin{figure*}[tb]
\centering
\begin{subfigure}[t]{0.25\textwidth}
\centering
\vskip 0pt 
\begin{tikzpicture}[->,>=stealth',shorten >=1pt,auto,node distance=1.5cm,thick]
  \node[unobserved node] (Z) {Z};
  \node[observed node] (T) [above right of=Z] {T};
  \node[observed node] (F) [below right of=Z] {F};

  \path
    (Z) edge (F)
    (Z) edge (T);
\end{tikzpicture}
\caption{Basic NMF (Section~\ref{sec:nmfderivation})}
\label{fig:zft}
\end{subfigure}
~
\begin{subfigure}[t]{0.35\textwidth}
\centering
\vskip 0pt 
\begin{tikzpicture}[->,>=stealth',shorten >=1pt,auto,node distance=1.5cm,thick]
  \node[unobserved node] (S) {S};
  \node[unobserved node] (Z) [right of=S] {Z};
  \node[observed node] (T) [above right of=Z] {T};
  \node[observed node] (F) [below right of=Z] {F};

  \path
    (S) edge (Z)
    (S) edge (F)
    (S) edge (T)
    (Z) edge (F)
    (Z) edge (T);
\end{tikzpicture}
\caption{NMF for source separation (Section~\ref{sec:nmfsourcesep})}
\label{fig:szft}
\end{subfigure}
~
\begin{subfigure}[t]{0.35\textwidth}
\centering
\vskip 0pt 
\begin{tikzpicture}[->,>=stealth',shorten >=1pt,auto,node distance=1.5cm,thick]
  \node[unobserved node] (S) {S};
  \node[observed node] (D) [below of=S] {D};
  \node[unobserved node] (Z) [right of=S] {Z};
  \node[observed node] (T) [above right of=Z] {T};
  \node[observed node] (F) [below right of=Z] {F};

  \path
    (S) edge (D)
    (S) edge (Z)
    (S) edge (F)
    (S) edge (T)
    (Z) edge (F)
    (Z) edge (T);
\end{tikzpicture}
\caption{NTF with directionality (Section~\ref{sec:ntfmodel})}
\label{fig:szftd}
\end{subfigure}
\caption{Graphical models for the factorizations in this paper.  The input data is a joint distribution over the shaded variables.}
\end{figure*}

We use Minorization-Maximization, a generalization of EM \cite{hl:qrvmma}, to iteratively optimize over $q$.  Fix a factored distribution $q^0$.  The essential step is to find a factored distribution $q^1$ with higher cross entropy. Jensen's inequality on the logarithm gives:
\begin{equation*}
 \sum_z q^0(z\mid f,t)\log \frac{q(f,t,z)}{q^0(z\mid f,t)}
 \leq \log\sum_z q^0(z\mid f,t)\frac{q(f,t,z)}{q^0(z\mid f,t)} = \log\sum_z q(f,t,z)
\end{equation*}
for all $f,t,$ and $q$.  When $q = q^0$ we have equality, since all the terms being averaged are equal.  Substituting into $\alpha(q)$ gives 
\begin{equation*}
\beta(q) \eqdef \sum_{f,t,z} \pobs(f,t) q^0(z\mid f,t) \log \frac{q(f,t,z)}{q^0(z\mid f,t)} \leq \alpha(q) ,
\end{equation*}
again with equality at $q = q^0$ ($\beta$ is said to \textbf{minorize} $\alpha$ at $q^0$).  If $q^1$ maximizes $\beta$ then $\alpha(q^1)\geq \beta(q^1) \geq \beta(q^0) = \alpha(q^0)$.

The denominator in $\beta$ only contributes an additive constant, so we can equivalently maximize the cross entropy $\gamma$ between $r(f,t,z)\eqdef \pobs(f,t) q^0(z\mid f,t)$ and $q(f,t,z)$:
\begin{equation*}
\gamma(q) \eqdef \sum_{f,t,z} r(f,t,z) \log q(f,t,z)
= \sum_z r(z) \sum_f r(f\mid z) \log q(f\mid z) + \sum_{t,z} r(t, z) \log q(t, z).
\end{equation*}
Though our original goal was to maximize a cross entropy, we have made progress in the senses that (a) there is no longer a sum inside the logarithm and (b) we have decoupled the terms involving $q(f\mid z)$ and $q(t, z)$.  These can be chosen independently and arbitrarily while maintaining the factored form $q(f,t,z)=q(f\mid z)q(t, z)$.

To find the optimal $q^1$ we apply Gibbs' inequality: for a fixed probability distribution $\sigma$, the probability distribution $\tau$ which maximizes the cross entropy $\sum_u \sigma(u)\log \tau(u)$ is $\tau = \sigma$.  Therefore we maximize $\gamma$ by choosing $q^1(f\mid z) \eqdef r(f\mid z)$ and $q^1(t, z) \eqdef r(t, z)$.  Typically $q^1(f,t,z)\neq r(f,t,z)$; rather $q^1(f,t,z)$ is a product of a marginal and a conditional of $r$, which itself might not factor.

These updates can be viewed as alternating projections.  We seek a distribution $q(f,t,z)$ which (a) factors and (b) has marginal $q(f,t)$ close to $\pobs(f,t)$.  We begin with $q^0(f,t,z)$ which satisfies (a) but not (b) and modify it to get $r(f,t,z) = \pobs(f,t)q^0(z\mid f,t)$, which gives (b) exactly but destroys (a).  Then we multiply a marginal and conditional of $r(f,t,z)$ to get $q^1(f,t,z)$, sacrificing (b) to satisfy (a), and repeat.

\subsection{Multiplicative updates for NMF}
\label{sec:nmfmultupdate}
These iterations can be computed efficiently via the celebrated multiplicative updates  \cite{ls:anmf}.  To compute
\begin{align*}
q^1(t,z) &= \sum_f r(f,t,z) = \sum_f \pobs(f,t)q^0(z\mid f,t) = \sum_f \pobs(f,t)\frac{q^0(f\mid z)q^0(t, z)}{q^0(f,t)}\\
& = q^0(t,z)\sum_f \underbrace{\frac{\pobs(f,t)}{q^0(f,t)}}_{\text{call this }\rho(f,t)}q^0(f\mid z) = q^0(t,z)\sum_f \rho(f,t)q^0(f\mid z),
\end{align*}
matrix multiply to find $q^0(f,t) = \sum_{z'} q^0(f\mid z')q^0(t, z')$, elementwise divide to get $\rho(f,t) \eqdef \pobs(f,t)/q^0(f,t)$, matrix multiply for $\sum_f \rho(f,t)q^0(f\mid z)$, and elementwise multiply by $q^0(t,z)$.  Reuse $\rho(f,t)$ to compute $q^1(f,z)$ analogously and condition to get $q^1(f\mid z)$.

This method avoids storing any $F\times T\times Z$ arrays, such as $r(f,t,z)$ or $q^0(z\mid f,t)$.  Indeed, this implementation uses $\Theta(FT + FZ + TZ)$ memory (proportional to output plus input) total, but $\Theta(FTZ)$ arithmetic operations per iteration.

\subsection{Supervised NMF for single-channel audio source separation}
\label{sec:nmfsourcesep}
Though our focus is on unsupervised methods, we recall for comparison how to use NMF to separate $S$ audio sources.  We decompose the mixture as a weighted sum over sources
$s$ per Figure~\ref{fig:szft}:
\begin{equation}
\label{eq:nmfsep}
\pobs(f,t) \approx \sum_s q(s)\sum_z q(f\mid z,s) q(t,z \mid s).
\end{equation}
Mathematically, this is NMF with inner dimension $ZS$.  Fitting such a model as in Section~\ref{sec:nmfderivation} gives no separation: the cross entropy objective is invariant to swapping dictionary elements between sources, so it does not encourage all dictionary elements of a modeled source to correspond to a single physical source.

A typical workaround uses training data $\phatobs_s(f,t)$ corresponding to recordings of sounds typical of each of the sources alone.  We apply NMF to these for each $s$ separately, learning a representative dictionary $\hat{q}_s(f\mid z)$.  The factor $\hat{q}_s(t,z)$ represents when and how active these dictionary elements are in the training data, so it is discarded as irrelevant for separation (this does lose information about time evolution of activations).

We apply NMF again to learn \eqref{eq:nmfsep} with the twist that we fix $q(f\mid z,s)\eqdef \hat{q}_s(f\mid z)$ for all iterations, so the model cannot freely swap dictionary elements between sources.  Instead of improving both terms in $\gamma(q)$ we improve one and leave the other fixed; the cross entropy still increases. After NMF, $q(s\mid f,t)$ gives a measure of the contribution from source $s$ in each time-frequency bin.

A common use case is when $\pobs(f,t)$ is a normalized spectrogram, computed as the magnitude of a Short-Time Fourier Transform (STFT).  It is typical to approximately reconstruct the time-domain audio of a separated source $s$ by multiplying the magnitude component $\pobs(f,t)q(s\mid f,t)$ with the phase of the mixture STFT, then taking the inverse STFT.  Considering the outputs to be the mask $q(s\mid f,t)$ or reconstructed time-domain audio, separation takes $\Theta(FTS + FZS + TZS)$ memory total and $\Theta(FTZS)$ arithmetic operations per iteration.

\subsection{Direction of Arrival (DOA) estimation}
\label{sec:doa}
Estimating DOA at an array is a well-studied problem \cite{bch:masp} and various methods can be used in the framework of Section~\ref{sec:ntf}.  We use the \textbf{least squares method} (perhaps the simplest) for estimating a DOA at each time-frequency bin \cite{t:msstpdrsc}.  We take as given the STFTs of audio signals recorded at each of $M$ microphones.  The same procedure is applied to all bins, so we focus on a single bin and its STFT values $Y_1, \ldots, Y_M$.

Assume this bin is dominated by a point source far enough away to appear as a plane wave and the array is small enough to avoid wrapping the phases $\angle Y_i$.  Letting $x_i$ denote the position of microphone $i$ and $k$ the wave vector, we have $\angle Y_i - \angle Y_1 = (x_i-x_1)\cdot k$.  We solve these linear equations for $k$ in a least squares sense.  The direction of $k$ serves as a DOA estimate for the chosen bin.  The coefficients of $k$ are fixed by the geometry, so the least squares problems for all time-frequency bins can be solved with a single pseudoinverse at design time and a small matrix multiplication for each bin thereafter.

\section{Nonnegative Tensor Factorization (NTF): Adding Directionality to NMF}
\label{sec:ntf}
\subsection{The Main Idea}
\label{sec:ntfmodel}
This section is parallel to Section~\ref{sec:nmfderivation}, except instead of a matrix $\pobs(f,t)$ we take as given an array or tensor $\pobs(f,t,d)$, interpreted as a distribution over time, frequency, and DOA quantized to a finite domain of size $D$.  In practice we take $\pobs(f,t,d) = \pobs(f,t)\pobs(d\mid f,t)$, where $\pobs(f,t)$ is again a normalized spectrogram and $\pobs(d\mid f,t)$ is an estimate of direction per time-frequency bin.  There is little amplitude variation between the closely-spaced microphones, so we can derive $\pobs(f,t)$ from any of them -- spatial diversity is captured by $\pobs(d\mid f,t)$.  For efficiency we choose $\pobs(d\mid f,t)$ to place all weight on the DOA estimated as in Section~\ref{sec:doa}, but we could also use e.g.\ the normalized output of a family of $D$ beamformers as in \cite{tsws:dnmfjsls}.

We fit $\pobs(f,t,d)\approx\sum_{s,z}q(f,t,d,z,s)$ for the factorization
\begin{equation}
\label{eq:ntfsep}
q(f,t,d,z,s) \eqdef q(s)q(f\mid s,z)q(t,z\mid s)q(d\mid s) = q(d,s)q(f\mid s,z)q(t,z\mid s),
\end{equation}
represented in Figure~\ref{fig:szftd}.  A distribution $q(d\mid s)$ rather than a fixed DOA per source allows for noise, slight movements of sources, and modeling error.

Crucially, \eqref{eq:ntfsep} forces $q(d\mid s, z) = q(d \mid s)$ not to depend on $z$: dictionary elements corresponding to the same source explain energy coming from the same direction.  In particular, cross entropy on this model is not invariant to swapping dictionary elements between sources.  If the model tried to account for multiple physical sources within a single source $s$ by choosing a multimodal $q(d\mid s)$, the cross entropy would be low because some dictionary elements for $s$ would not have energy at some modes.  We can thus hope to learn the model \eqref{eq:ntfsep} from $\pobs(f,t,d)$ alone, without training data.  

We use Minorization-Maximization to fit \eqref{eq:ntfsep} to $\pobs(f,t,d)$, just as we did to fit \eqref{eq:nmfdecomp} to $\pobs(f,t)$ in Section~\ref{sec:nmfderivation}.  The same argument leads us to begin with a factored model
\begin{equation*}
q^0(f,t,d,z,s) \eqdef q^0(d,s)q^0(f\mid s,z)q^0(t,z\mid s),
\end{equation*}
force the desired marginal to obtain
\begin{equation*}
r(f,t,d,z,s) \eqdef \pobs(f,t,d)q^0(z,s\mid f,t,d),
\end{equation*}
and return to factored form by computing conditionals of $r$:
\begin{equation*}
{q^1(d,s) \eqdef r(d,s)},\qquad {q^1(f\mid s,z) \eqdef r(f \mid s,z)},\qquad q^1(t,z\mid s) = r(t,z\mid s).
\end{equation*}

We iterate, then compute the soft mask $q(s\mid f,t)$ as in Section~\ref{sec:nmfsourcesep}.  Without training data the correspondence between learned sources and sources in the environment is unknown a priori, but the learned factors $q(d\mid s)$ can help disambiguate.

\subsection{Implementation}
\label{sec:ntfmultupdate}

As in Section~\ref{sec:nmfmultupdate} we can turn these equations into multiplicative updates and in the process reduce the resource requirements.  The savings come from ordering and factoring the operations appropriately as well as expressing tensor operations in terms of matrix multiplications (of course to multiply matrices $A$ and $B$ we would not waste memory computing the full three-dimensional tensor $A_{ij}B_{jk}$ before summing; when possible we avoid this for tensors as well). For example, we can calculate
\begin{equation}
\label{eq:ntfmultupdate}
\begin{split}
q^1(d,s) 
& = \sum_{f,t,z} r(f,t,d,z,s)
= q^0(d,s)\sum_{f,t} \frac{\pobs(f,t,d)}{\sum_{s'} q^0(d,s')q^0(f,t \mid s')}q^0(f,t \mid s) \\
& = q^0(d,s)\sum_{f,t} \underbrace{\frac{\pobs(f,t,d)}{q^0(f,t,d)}}_{\text{call this }\rho(f,t,d)}q^0(f,t \mid s)
= q^0(d,s)\sum_{f,t} \rho(f,t,d)q^0(f,t \mid s),
\end{split}
\end{equation}
as follows.  Compute $q^0(f,t\mid s)$; for each $s$ this is an $F\times Z$ times $Z\times T$ matrix multiplication.  Then compute the denominator $q^0(f,t,d)$ as a $D\times S$ times $S\times FT$ matrix multiplication.  Divide $\pobs(f,t,d)$ elementwise by the result and call this $\rho(f,t,d)$.  Compute the remaining sum as a $D\times FT$ times $FT\times S$ matrix multiplication.  Multiply by $q^0(d,s)$ elementwise to get $q^1(d,s)$.

Reusing $\rho$, similar computations yield $q^1(f\mid z,s)$ and $q^1(t,z\mid s)$.  The total memory required is $\Theta(FTS+FTD+FZS+TZS)$, again proportional to the memory required to store the input and output (both the factorization and the mask $q(s\mid f,t)$ are here considered part of the output).  The number of arithmetic operations used at each iteration is $\Theta(FTS(D+Z))$.  So in addition to never having to allocate memory for any of the size $FTDZS$ arrays referred to in Section~\ref{sec:ntfmodel}, we do not even have to explicitly compute all their elements individually.

\subsection{Sparse Direction Data}
\label{sec:sparsedir}
Suppose all the mass in each time-frequency bin is assigned to a single direction $d(f,t)$ (e.g.\ the output of Section~\ref{sec:doa}, discretized), so $\pobs(d \mid f,t) = \delta(d = d(f,t))$ in terms of the Kronecker $\delta$.  The input ($\pobs(f,t)$ and $d(f,t)$) then has size $\Theta(FT)$ and the implementation simplifies further.

Since $r(f,t,d,z,s)$ is only nonzero when $d = d(f,t)$, we only need to compute the denominator $q^0(f,t,d)$ of \eqref{eq:ntfmultupdate} for $d=d(f,t)$.  To do this, we compute $q^0(f,t\mid s)$ as before and then evaluate $q^0(d,s)$ at $d = d(f,t)$, yielding another $F\times T\times S$ tensor $q^0(d(f,t),s)$.  Summing the elementwise product $q^0(f,t\mid s')q^0(d(f,t),s')$ over $s'$ yields the $F\times T$ array $q^0(f,t,d(f,t))$.

Instead of defining $\rho(f,t,d)$ as in Section~\ref{sec:ntfmultupdate} we define $\rho(f,t) \eqdef \frac{\pobs(f,t)}{q^0(f,t,d(f,t))}$, so
\begin{equation*}
r(f,t,d,z,s) = \rho(f,t)\delta(d = d(f,t))q^0(d,s)q^0(f\mid z,s)q^0(t,z\mid s).
\end{equation*}
Marginalizing, we get:
\begin{equation*}
q^1(d,s) = \sum_{f,t,z} r(f,t,d,z,s) = q^0(d,s)\sum_{f,t\,:\,d(f,t) = d} \rho(f,t)q^0(f,t \mid s),
\end{equation*} 
which can now be computed naively in $\Theta(FTS)$ memory and arithmetic operations.  For
\begin{equation*}
q^1(f,z,s) = \sum_{t,d} r(f,t,d,z,s) = q^0(f\mid z,s) \sum_t \rho(f,t)q^0(d(f,t),s) q^0(t,z\mid s)
\end{equation*}
we multiply $\rho(f,t)$ by $q^0(d(f,t),s)$, which takes $\Theta(FTS)$ memory and operations, then compute the sum over $t$ as $S$ matrix multiplications of size $F\times T$ times $T\times Z$.  This takes $\Theta(FZS)$ memory and $\Theta(FTZS)$ operations.  Then we multiply elementwise by $q^0(f\mid z,s)$ and condition to get $q^1(f\mid z,s)$.  The computation of $q^1(t,z\mid s)$ is similar.

The resource requirements are $\Theta(FTS + FZS + TZS)$ memory total and $\Theta(FTZS)$ arithmetic operations per iteration\footnote{Strictly speaking $D$ must enter somewhere.  We assume the expected use case in which $D<FT$.}.  This is the same order as supervised NMF (Section~\ref{sec:nmfsourcesep}), though these are apples and oranges: one uses direction information and the other uses clean audio training data.

\subsection{Capturing Geometry by Constraining $q(d\mid s)=q_{\theta_s}(d\mid s)$}
\label{sec:geo}
So far the algorithm is invariant to permuting the direction labels $d$: we have not told it if $d=1$ is close to $d=2$, etc. In favorable circumstances, when there is low noise and EM does not get stuck in bad local optima, we experimentally observe that the algorithm infers the geometry from the data in the sense that the learned $q(d\mid s)$ varies smoothly with $d$ for each $s$.  In less favorable circumstances an underlying source of sound can be split between multiple ``separated'' sources reported by the algorithm, resulting in audible artifacts and qualitatively poor separation.

We improve output quality by enforcing structure on $q(d\mid s)$.  For concreteness we consider the case when the directions $d=1,\ldots D$ indicate azimuthally equally spaced angles all the way around the microphone array.  We constrain $q(d\mid s)$ to the two-dimensional exponential family with sufficient statistics $\phi(d) = \left(\sin(2\pi d/D), \cos(2\pi d/D)\right)$, a discretized version of the von Mises family.  The distributions $q_{\theta_s}(d\mid s) \propto \exp(\phi(d)\theta_s)$ look like unimodal bumps on a (discretized) circle, with the polar coordinates of the natural parameters $\theta_s$ controlling the position and height of the bump.

At a particular iteration call the exact maximizer of the minorizer (with no exponential family constraint) given by Gibbs' inequality $\hat{q}(d\mid s)$; we saw how to compute this in Sections~\ref{sec:ntfmultupdate}~and~\ref{sec:sparsedir}.  Replacing $q(d\mid s)$ in the cross entropy objective with $q_{\theta_s}(d\mid s)$, the same argument as before goes through and all distributions except $q_{\theta_s}(d\mid s)$ are computed in the same way.  The term involving $q_{\theta_s}(d\mid s)$ works out to be $\sum_d \hat{q}(d\mid s) \log q_{\theta_s}(d\mid s)$ and its gradient with respect to $\theta_s$ is $\sum_d \left[q_{\theta_s}(d\mid s) - \hat{q}(d\mid s)\right]\phi(d)$, the difference in the moments of $\phi(d)$ with respect to $q_{\theta_s}$ and the target $\hat{q}$.  Since the updates are being iterated anyway, we observe empirically that it suffices to fix a step size $\lambda$ and take a single gradient step in $\theta_s$ at each iteration:
\begin{equation*}
\theta_s^{\text{new}} = \theta_s^{\text{old}} + \lambda \sum_d\left[q_{\theta_s^{\text{old}}}(d\mid s) - \hat{q}(d\mid s)\right]\phi(d).
\end{equation*}

As shown in Table~\ref{tab:results} this change increases output quality significantly with only a marginal effect on run time.

\section{Experiments}
\label{sec:experiments}
\begin{table}[tbp]
\centering
\begin{tabular}{lrrrr}
\toprule
&\multicolumn{3}{c}{\texttt{BSS\_EVAL} in dB} & \multicolumn{1}{c}{run time as} \\
Algorithm & SDR & SIR & SAR & \multicolumn{1}{c}{\% real time}\\
\midrule
Ideal Binary Mask & 13.9 & 22.7 & 14.7 & 1.4 \% \\
Ideal Ratio Mask & 13.1 & 18.6 & 15.0 & 1.5 \% \\
\midrule
\textbf{Directional NTF, constrained $q_{\theta_s}(d\mid s)$} & 9.6 & 14.6 & 14.2 & 20.8 \% \\
Directional NTF, unconstrainted $q(d\mid s)$ & 5.6 & 10.4 & 12.2 & 20.7 \% \\
Directional NMF & 3.0 & 6.8 & 10.4 & 13.7 \% \\
Supervised NMF & 2.2 & 4.5 & 10.3 & 10.4 \% \\

\bottomrule
\end{tabular}
\caption{Results of Section~\ref{sec:experiments} experiments averaged over $1000$ random instances.  Ideal masks use ground truth and should be viewed as bounds only.  A range of $\pm 0.5$ is an (at least) $95\%$ confidence interval for all true average \texttt{BSS\_EVAL} metrics.  Runtimes are from python code (\codeurl) on a $2015$ Macbook Pro.}
\label{tab:results}
\end{table}

\begin{figure}
\includegraphics[width=\textwidth]{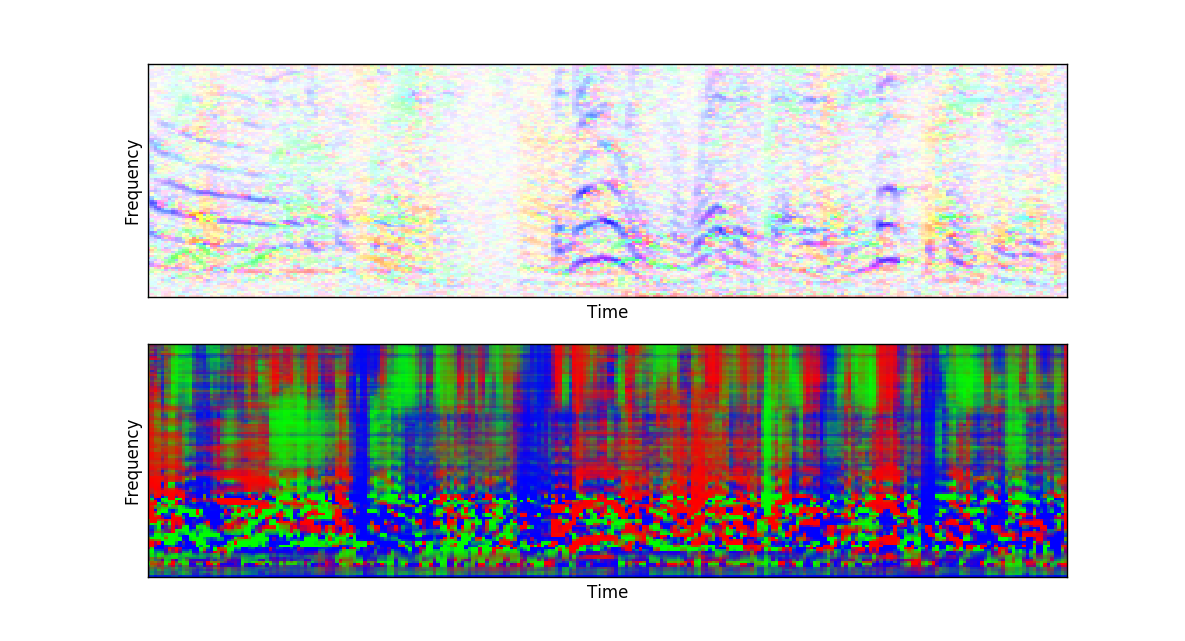}
\caption{The top shows a portion of a spectrogram from a real recording of three people talking simultaneously, each about $1$ meter from a $3\text{mm}\times 5\text{mm}$ rectangular microphone array, with time-frequency bins colored according to azimuth angle estimated as in Section~\ref{sec:doa}.  Interpreting color as height out of the page, this is the sparse time $\times$ frequency $\times$ direction tensor used as the input to (Sparse) Directional NTF.  The bottom shows the computed masks for separating this signal into three sources with $q(s\mid f, t)$ for $s=1,2,3$ being interpreted as the red, blue, and green channels, respectively.  Raw input and separated output audio are available in the supplemental materials.}
\label{fig:ntf_example}
\end{figure}

Qualitatively, Directional NTF separates well on a $3\text{mm}\times 5\text{mm}$ rectangular MEMS microphone array we built.  Audio of three people talking simultaneously recorded with that array and the corresponding separated output is available in the supplemental materials.  A portion of this recording and its separation into sources is shown in Figure~\ref{fig:ntf_example}.

For a quantitative experiment, we simulated this array in a $3\text{m}\times 4\text{m}$ room.  For each of $1000$ instances of the experiment, we randomly selected two TIMIT sentences \cite{timit} from different speakers in the TIMIT test data set and used the code from \cite{dsv:rcp} to simulate these sentences being spoken simultaneously in the room.  We simulated data due to the lack of publicly available data with such closely-spaced microphones and to enable us to use the ground truth to quantify performance.  Source and array locations were uniformly random, conditioned to be at least $0.5\text{m}$ from the walls.

We compare four factorization-based source-separation algorithms: Directional NTF as in Section~\ref{sec:sparsedir} with the constrained $q_{\theta_s}(d\mid s)$ from Section~\ref{sec:geo}; Directional NTF with unconstrained $q(d\mid s)$; a less structured version called Directional NMF \cite{tsws:dnmfjsls}, which consists of factoring $\pobs(f,t,d) \approx \sum_s q(f,t \mid s) q(d,s)$, to highlight the importance of imposing structure on the separated sources; and Supervised NMF \cite{rs:lvdsscss} as reviewed in Section~\ref{sec:nmfsourcesep}.  The first three methods receive only the four-channel mixed audio, while the fourth receives one channel of mixed audio and a different clean TIMIT sentence of training data for each speaker.  For an upper bound we compare two oracular masks.  See Table~\ref{tab:results} for results using the \texttt{mir\_eval} implementation \cite{mir_eval} of the \texttt{BSS\_EVAL} metrics \cite{bss_eval}.

All algorithms are set to extract $S=2$ sources.  Directional NTF and Supervised NMF each model sources with $Z = 20$ dictionary elements; Directional NMF has no such parameter.  Directional methods receive a least-squares estimated azimuthal DOA angle for each time-frequency bin (Section~\ref{sec:doa}) quantized to $D=24$ levels and the constrained $q_{\theta_s}(d\mid s)$ method uses a learning rate $\lambda=2$ for the natural parameters.  Note that in \cite{tsws:dnmfjsls} the Directional NMF model was used with a dense estimate of energy as a function of direction, rather than a sparse estimate of a single dominant direction for each time-frequency bin, so these results are not directly comparable to that paper.

\section{Conclusions}
\label{sec:conclusions}

Directional NTF is better than the other (non-oracular) algorithms according to all \texttt{BSS\_EVAL} metrics (Table~\ref{tab:results}).  Real mixtures admit qualitative perceptual improvements in line with simulation.  We close with directions for future work.

First, this method fits naturally into the basic NMF / NTF framework.  As such it should be extensible using the many improvements to these methods available in the literature.

Second, when blindly separating speech from background noise, it is an open question how to automatically determine which source is speech.  In some applications one can infer this from the centers of the learned distributions $q(d\mid s)$ and prior information about the location of the speaker.  In others one may expect diffuse noise and call the source with $q(d \mid s)$ more tightly-peaked the speaker.  A more broadly-applicable source selection method would be desirable.

Third, we leave analysis of performance as a function of geometry and level of reverberation for future work.

\ack

\small

\bibliographystyle{plain}
\bibliography{LyricReferences}

\end{document}